\documentclass[11pt,a4paper]{article}
\usepackage[hyperref]{acl2020}
\usepackage{times}
\usepackage{latexsym}

\usepackage{graphicx}  
\usepackage{amsmath}
\usepackage{algorithm}
\usepackage{algpseudocode}
\usepackage{amsfonts}
\usepackage{amssymb}
\usepackage{bm}
\usepackage{color}
\usepackage{microtype}
\usepackage{pgfplots}
\usepackage{subcaption}
\usepackage{xcolor}
\usepackage{multirow}
\usepackage{enumitem}

\pgfplotsset{compat=1.12}

\usepackage{relsize}

\newenvironment{enumeratesquish}{\begin{list}{\addtocounter{enumi}{1}\labelenumi}{\setlength{\itemsep}{0em}\setlength{\labelwidth}{0.75em}\setlength{\leftmargin}{\labelwidth}\addtolength{\leftmargin}{\labelsep}}}{\end{list}\setcounter{enumi}{0}}

\aclfinalcopy 

\title{Contextual Neural Machine Translation Improves Translation of Cataphoric Pronouns}

\author{KayYen Wong$^\dagger$ \\
    \And
  Sameen Maruf$^\ddagger$ \\
  Faculty of Information Technology, Monash University, VIC, Australia \\
  {$^\dagger$\texttt{kywon63@student.monash.edu}} \\
  {$^\ddagger$\texttt{firstname.lastname@monash.edu}}\\
   \\\And
  Gholamreza Haffari$^\ddagger$ \\
  }

\date{}

\begin{document}
\maketitle
\begin{abstract}
The advent of context-aware NMT has resulted in promising improvements in the overall translation quality and specifically in the translation of discourse phenomena such as pronouns. Previous works have mainly focused on the use of past sentences as context with a focus on anaphora translation. In this work, we investigate the effect of future sentences as context by comparing the performance of a contextual NMT model trained with the future context to the one trained with the past context. Our experiments and evaluation, using generic and pronoun-focused automatic metrics, show that the use of future context not only achieves significant improvements over the context-agnostic Transformer, but also demonstrates comparable and in some cases improved performance over its counterpart trained on past context. We also perform an evaluation on a targeted cataphora test suite and report significant gains over the {context-agnostic} Transformer in terms of BLEU.
\end{abstract}

\section{Introduction}
Standard machine translation (MT) systems typically translate sentences in isolation, ignoring essential contextual information, where a word in a sentence may reference other ideas or expressions within a piece of text. This locality assumption hinders the accurate translation of referential pronouns, which rely on surrounding contextual information to resolve cross-sentence references. The issue is further exacerbated by 
differences in pronoun rules between source and target languages, often resulting in morphological disagreement in the quantity and gender of the subject being referred to  \cite{Vanmassenhove:18}.

Rapid improvements in NMT have led to it replacing SMT as the dominant paradigm. With this, context-dependent NMT has gained traction, overcoming the locality assumption in SMT through the use of additional contextual information.
This has led to improvements in not only the overall translation quality but also pronoun translation \cite{Jean:17, Bawden:17, Voita:18, Miculicich:18}. However, all these works have neglected the context from \emph{future} sentences, with \newcite{Voita:18} reporting it to have a negative effect on the overall translation quality.

In this work, we investigate the effect of future context in improving NMT performance. We particularly focus on pronouns and analyse corpora from different domains to discern if the future context could actually aid in their resolution. We find that for the Subtitles domain roughly 16\% of the pronouns are cataphoric. This finding motivates us to investigate the performance of a context-dependent NMT model \cite{Miculicich:18} trained on the future context in comparison to its counterpart trained on the past context. We evaluate our models in terms of overall translation quality (BLEU) and also employ three types of automatic pronoun-targeted evaluation metrics. We demonstrate strong improvements for all metrics, with the model using future context showing comparable or in some cases even better performance than the one using only past context. We also extract a targeted cataphora test set and report significant gains on it with the future context model over the baseline.

\section{Related Work}\label{sec:lit}

\paragraph{Pronoun-focused SMT} Early work in the translation of pronouns in SMT attempted to exploit coreference links as additional context to improve the translation of anaphoric pronouns (\citealt{Le-Nagard:10}; \citealt{Hardmeier:10}). These works yielded mixed results which were attributed to the limitations of the coreference resolution systems used in the process \cite{Guillou:12}.

\paragraph{Context-Aware NMT} Multiple works have successfully demonstrated the advantages of using larger context in NMT, where the context comprises few previous source sentences \cite{Wang:17, Zhang:18}, few previous source and target sentences \cite{Miculicich:18}, or both past and future source and target sentences \cite{Maruf:18, Maruf2018, Maruf:2019}.

\setlength{\tabcolsep}{3pt}
\begin{table}[t]
\centering
{\small
\begin{tabular}{l||c|c}
\textbf{Domain} & 
\textbf{\#Sentences} & \textbf{Document length}\\
\hline \hline
& \multicolumn{2}{c}{\textbf{English-German}}\\
\cline{2-3}
{Subtitles} & 
9.39M/9K/14.1K & 565.8/582.2/591.0 \\
Europarl & 
1.67M/3.6K/5.1K & 14.1/15.0/14.1 \\
TED Talks & 
0.21M/9K/2.3K & 120.9/96.4/98.7\\
\hline \hline
& \multicolumn{2}{c}{\textbf{English-Portuguese}}\\
\cline{2-3}
{Subtitles} & 
15.2M/16.1K/23.6K & 580.4/620.6/605.3 \\
\end{tabular}
}
\caption{Train/dev/test statistics: number of sentences (K: thousands, M: millions), and average document length (in sentences). The \textit{\#Documents} can be obtained by dividing the \textit{\#Sentences} by the \textit{Document Length}.}
\label{table:corpora}
\vspace{-4mm}
\end{table}

Further, context-aware NMT has demonstrated improvements in pronoun translation using past context, through concatenating source sentences \cite{Tiedemann:17} or through an additional context encoder \cite{Jean:17, Bawden:17, Voita:18}. \citet{Miculicich:18} observed reasonable improvements in generic and pronoun-focused translation using three previous sentences as context. \citet{Voita:18} observed improvements using the previous sentence as context, but report decreased BLEU when using the following sentence. We, on the other hand, observe significant gains in BLEU when using the following sentence as context on the same data domain.


\section{Contextual Analysis of Corpora}\label{sec:data}

To motivate our use of the future context for improving the translation of cataphoric pronouns in particular and NMT in general, we first analyse the distribution of coreferences for anaphoric and cataphoric pronouns over three different corpora - 
OpenSubtitles2018\footnote{\url{http://www.opensubtitles.org/}
} \cite{Lison:16}, Europarl \cite{Koehn:05} and TED Talks \cite{Cettolo:12} - for English-German. For Europarl and TED Talks, we use the publicly available document-aligned version of the corpora \cite{Maruf:2019}. For Subtitles, we align the English and German subtitles at the document-level using publicly available alignment links.\footnote{\url{http://opus.nlpl.eu/OpenSubtitles2018.php}} To control for the length and coherency of documents, we keep subtitles with a run-time less than 50 minutes (for English) and those with number of sentences in the hundreds. 
The corpus is then randomly split into training, development and test sets in the ratio 100:1:1.5.  Table~\ref{table:corpora} presents the corpora statistics.

\setlength{\tabcolsep}{2.5pt}
\begin{table}[t]
\centering
{\small
\begin{tabular}{c||c|c|c}
\textbf{Pronoun} & 
\textbf{Subtitles} & \textbf{Europarl} & \textbf{TED Talks}\\
\hline \hline
Intrasentential & 30.1 & 75.6 & 64.1 \\
Anaphora ($<$ 0) & 54.3 & 19.6 & 28.5 \\
Cataphora ($>$ 0) & 15.6 & 4.7 & 7.4 \\
\end{tabular}
}
\caption{Percentage of different pronoun types.}
\label{table:corporastats}
\vspace{-2mm}
\end{table}

\begin{figure}[t]
\centering
{
\begin{tikzpicture}
    \begin{axis}[
            ybar stacked,
            x=0.4cm,
            bar width=0.21cm,
            height=.53\linewidth,
            legend style={at={(0.64,1)},
                anchor=north,legend columns=-1},
            symbolic x coords={0, 1, 2, 3, 4, 5, 6, 7, 8, 9, 10},
            xtick={0, 1, 2, 3, 4, 5, 6, 7, 8, 9, 10},
	every node near coord/.append style={font=\scriptsize},
            ymin=0,ymax=40,
            ylabel={\%age of occurences},
        ]
        \addplot[fill=violet!60] coordinates{
        (1, 30.4) (2, 8.5) (3, 4.1) (4, 2.5) (5, 1.6) (6, 1.2) (7, 0.8) (8, 0.7) (9, 0.5) (10, 0.4)};
        
        \addplot[fill=orange!80] coordinates{
        (1, 5.6) (2, 2.2) (3, 1.2) (4, 0.8) (5, 0.6) (6, 0.5) (7, 0.4) (8, 0.3) (9, 0.3) (10, 0.2)};
        \legend{\scriptsize{Anaphora}, \scriptsize{Cataphora}}
    \end{axis}
\end{tikzpicture}
}
\caption{Plot showing proportion of intersentential English pronouns versus size of coreference resolution window for the Subtitles corpus (plots for Europarl and TED Talks are in the appendix
).}\label{fig:corpusanalysis}
\vspace{-6mm}
\end{figure}
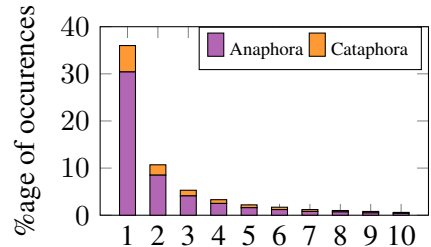

\paragraph{Analysis of Coreferences}
We find the smallest window within which a referential English pronoun is resolved by an antecedent or postcedent 
using \texttt{NeuralCoref}.\footnote{\url{https://github.com/huggingface/neuralcoref}} Table~\ref{table:corporastats} shows that the majority of pronouns in Europarl and TED Talks corpora are resolved intrasententially, while the Subtitles corpus demonstrates a greater proportion of intersentential coreferences. Further, anaphoric pronouns are much more frequent compared to cataphoric ones across all three corpora. For Subtitles, we also note that a good number of pronouns (15.6\%) are cataphoric, $\sim$37\% of which are resolved within the following sentence (Figure~\ref{fig:corpusanalysis}). This finding motivates us to investigate the performance of a context-aware NMT model (trained on Subtitles) for the translation of cataphoric pronouns.

\section{Experiments}
\paragraph{Datasets} 
We experiment with the Subtitles corpus on English-German and English-Portuguese language-pairs. 
To obtain English-Portuguese data, we employ the same pre-processing steps as reported in \S\ref{sec:data} (corpus statistics are in Table~\ref{table:corpora}). 
We use 80\% of the training data to train our models and the rest is held-out for further evaluation as discussed later in \S~\ref{subsec:pronouneval}.\footnote{Due to resource contraints, we use about two-thirds of the final training set ({$\sim$}8M sentence-pairs) for En-Pt.} The data is truecased using the Moses toolkit \cite{Koehn:07} and split into subword units using a joint BPE model with 30K merge operations \cite{Sennrich:16}.\footnote{Tokenisation is provided by the original corpus.} 


\setlength{\tabcolsep}{1.5pt}
\begin{table}[t]
\begin{center}
{\small
\begin{tabular}{l||c| c c}
\textbf{Lang. Pair} & \textbf{Baseline} & \textbf{HAN(\textit{k} = {+1})} & \textbf{HAN(\textit{k} = {-1})} 
\\
\hline \hline
English$\rightarrow$German & 31.87 & \textbf{32.53} 
& 32.48\\
German$\rightarrow$English & 35.92 & \textbf{36.64}$^{\spadesuit}$ 
& 36.32\\
English$\rightarrow$Portuguese & 35.45 & {36.04} 
& \textbf{36.21}\\
Portuguese$\rightarrow$English & 39.34 & \textbf{39.96}$^{\spadesuit}$ 
& 39.69\\
\end{tabular}
}
\end{center}
\caption{BLEU for the Transformer baseline and Transformer-HAN with following sentence (\textit{k} = +1) and previous sentence (\textit{k} = -1). 
$\spadesuit$: Statistically significantly better than HAN (\textit{k} = -1). 
}
\label{table:mainresults}
\vspace{-2mm}
\end{table}


\paragraph{Description of the NMT systems}\label{subsec:sysnmt}
As our baseline, we use the \texttt{DyNet} \cite{dynet} implementation of Transformer  \cite{Vaswani:17}.\footnote{\url{https://github.com/duyvuleo/Transformer-DyNet}} 
For the context-dependent NMT model, we choose the Transformer-HAN encoder \cite{Miculicich:18}, which 
has demonstrated reasonable performance for anaphoric pronoun translation on Subtitles. We extend its \texttt{DyNet} implementation \cite{Maruf:2019} to a single context sentence.
\footnote{
Where in the original architecture, \textit{k} sentence-context vectors were 
summarised into a document-context vector, we omit this step when using only one sentence in context.}\footnote{The code and data are available at \url{https://github.com/sameenmaruf/acl2020-contextnmt-cataphora}.}
For training, Transformer-HAN is initialised with the baseline Transformer and then the parameters of the whole network are optimised in a second stage as in  \newcite{Miculicich:18} (details of model configuration are in Appendix~\ref{section:app-modelconfig}). For evaluation,  we compute BLEU \cite{Papineni:02} on tokenised truecased text and measure statistical significance with \textit{p} $<$ 0.005 \cite{Clark:11}.

\subsection{Results}



We consider two versions of the Transformer-HAN respectively trained with the following and previous source sentence as context. From Table~\ref{table:mainresults}, we note both context-dependent models to significantly outperform the Transformer across all language-pairs in terms of BLEU. Further, HAN (\textit{k} = +1) demonstrates statistically significant improvements over the HAN (\textit{k} = -1) when translating to English. These results are quite surprising as \newcite{Voita:18} report decreased translation quality in terms of BLEU when using the following sentence for English$\rightarrow$Russian Subtitles. To identify if this discrepancy is due to the language-pair or the model, we conduct experiments with English$\rightarrow$Russian in the same data setting as \newcite{Voita:18} and find that HAN (\textit{k} = +1) still significantly outperforms the Transformer and is comparable to HAN (\textit{k} = -1) (more details in Appendix~\ref{section:app-enru}).

\begin{table}[t]
\centering
{\small
\begin{tabular}{l||c |c c c |c}
& \multicolumn{5}{c}{\textbf{English$\rightarrow$German}} \\
\cline{2-6}
\textbf{Model} & \textbf{APT} & \textbf{Precision} & \textbf{Recall} & \textbf{F1-score} & \textbf{CRC}\\
\hline \hline
Baseline & 60.8 & 47.4 & 54.3 & 50.7 & -\\
\hline
\ \ \ \ +HAN(\textit{k} = {+1}) & 61.4 & \textbf{48.3} & 54.3 & \textbf{51.1} & -\\
\ \ \ \ +HAN(\textit{k} = {-1}) & \textbf{62.0} & 48.0 & \textbf{54.6} & \textbf{51.1} & -\\
\hline
& \multicolumn{5}{c}{\textbf{German$\rightarrow$English}}\\
\cline{2-6}
\textbf{Model} & \textbf{APT} & \textbf{Precision} & \textbf{Recall} & \textbf{F1-score} & \textbf{CRC}\\
\hline \hline
Baseline & 77.9 & 56.9 & 50.4 & 53.4 & 50.4\\
\hline
\ \ \ \ +HAN(\textit{k} = {+1}) & \textbf{78.3} & {57.9} & \textbf{50.6} & \textbf{54.0} & 50.9\\
\ \ \ \ +HAN(\textit{k} = {-1}) & \textbf{78.3} & \textbf{58.0} & {50.5} & \textbf{54.0} & \textbf{51.0}\\
\hline \hline
& \multicolumn{5}{c}{\textbf{English$\rightarrow$Portuguese}} \\
\cline{2-6}
\textbf{Model} & \textbf{APT} & \textbf{Precision} & \textbf{Recall} & \textbf{F1-score} & \textbf{CRC}\\
\hline \hline
{Baseline} & 46.4 & 54.8 & \textbf{56.0} & {55.4} & -\\
\hline
\ \ \ \ {+HAN(\textit{k} = {+1})} & 47.0 & {55.8} & {55.2} & {55.5} & - \\
\ \ \ \ {+HAN(\textit{k} = {-1})} & \textbf{47.3} & {\textbf{56.0}} & {55.4} & {\textbf{55.7}} & - \\

\hline 
& \multicolumn{5}{c}{\textbf{Portuguese$\rightarrow$English}}\\
\cline{2-6}
\textbf{Model} & \textbf{APT} & \textbf{Precision} & \textbf{Recall} & \textbf{F1-score} & \textbf{CRC}\\
\hline \hline
Baseline & 
{64.3} & 54.9 & 51.1 & 53.0 & 50.2\\
\hline
\ \ \ \ +HAN(\textit{k} = {+1}) & 
{\textbf{64.6}} & \textbf{55.7} & \textbf{51.5} & \textbf{53.5} & 50.9\\
\ \ \ \ +HAN(\textit{k} = {-1}) & 
{64.3} & {55.6} & 51.2 & {53.4} & \textbf{51.6}\\
\end{tabular}
}
\caption{Pronoun-focused evaluation on generic test set for models trained on different types of context.}
\label{table:proneval}
\end{table}

\subsection{Analysis}\label{subsec:pronouneval}
\paragraph{Pronoun-Focused Automatic Evaluation}
For the models in Table~\ref{table:mainresults}, we employ three types of pronoun-focused automatic evaluation: 

\begin{enumeratesquish}
\item {\textbf{Accuracy of Pronoun Translation (APT)} \cite{Miculicich:17}\footnote{https://github.com/idiap/APT}. This measures the degree of overlapping pronouns between the output and reference translations obtained via word-alignments.}
\item {\textbf{Precision, Recall and F1 scores}. We use a variation of AutoPRF \cite{Hardmeier:10} to calculate precision, recall and F1-scores. For each source pronoun, we compute the clipped count \cite{Papineni:02} of overlap between candidate and reference translations. To eliminate word alignment errors, we compare this overlap over the set of dictionary-matched target pronouns, in contrast to the set of target words aligned to a given source pronoun as done by AutoPRF and APT.}
\item {\textbf{Common Reference Context (CRC)} \cite{Jwalapuram:19}. In addition to the previous two measures which rely on computing pronoun overlap between the target and reference translation, we employ an ELMo-based \cite{Peters:2018} evaluation framework that distinguishes between a good and a bad translation via pairwise ranking  \cite{Jwalapuram:19}. We use the CRC setting of this metric which considers the same reference context (one previous and one next sentence) for both reference and system translations. However, this measure is limited to evaluation only on the English target-side.\footnote{{We use the same English pronoun list for all pronoun-focused metrics (provided by \newcite{Jwalapuram:19} at \url{https://github.com/ntunlp/eval-anaphora}). All pronoun sets used in our evaluation are provided in Appendix~\ref{section:app-pronsets}.}}}

\end{enumeratesquish}


\begin{table}[t]
\centering
{\small
\begin{tabular}{l|c| c c}
\textbf{Model} & \textbf{BLEU} & \textbf{APT} & \textbf{F1-score}\\
\hline \hline
Baseline (Transformer) & 31.87 & 61.6 & 49.1\\
\hline
\ \ \ \ +HAN(\textit{k} = $\emptyset$) & 32.30 & 61.6 & 49.1\\
\ \ \ \ +HAN(\textit{k} = {+1,+2}) & 32.56$^{\spadesuit}$ & \textbf{62.0} & 49.8\\
\ \ \ \ +HAN(\textit{k} = {-2,-1}) & 32.47 & 61.9 & 49.8\\
\ \ \ \ +HAN(\textit{k} = {-2,-1,+1,+2}) & \textbf{32.59}$^{\spadesuit}$ & \textbf{62.0} & \textbf{49.9}\\
\end{tabular}
}
\caption{Evaluation on English$\rightarrow$German generic test set for HAN trained with \textit{k} = \{-2,-1,+1,+2\} but decoded with varying context. $\spadesuit$: Statistically significantly better than HAN with no context (\textit{k} = $\emptyset$). 
}
\label{table:ablation}
\vspace{-2mm}
\end{table}

{The results using the aforementioned pronoun evaluation metrics are reported in Table~\ref{table:proneval}.} We observe improvements for all metrics with both HAN models in comparison to the baseline. Further, we observe that the HAN (\textit{k} = +1) is either comparable to or outperforms HAN (\textit{k} = -1) on APT and F1 for 
De$\rightarrow$En and Pt$\rightarrow$En, suggesting that for these cases, the use of following sentence as context is at least as beneficial as using the previous sentence. For En$\rightarrow$De, we note comparable performance for the HAN variants in terms of F1, while for En$\rightarrow$Pt, the past context appears to be more beneficial.\footnote{It should be noted that for Portuguese, adjectives and even verb forms can be marked by the gender of the noun and these are hard to account for in automatic pronoun-focused evaluations.} In terms of CRC, we note HAN (\textit{k} = -1) to be comparable to (De$\rightarrow$En) or better than HAN (\textit{k} = +1) (Pt$\rightarrow$En). We attribute this to the way the metric is trained to disambiguate pronoun translations based on only the previous context and thus may have a bias for such scenarios.

\begin{table}[t]
\centering
{\small
{\begin{tabular}{l||c| c c c}
& \multicolumn{4}{c}{\textbf{English$\rightarrow$German}} \\
\cline{2-5}
\textbf{Model} & \textbf{Cataphora} & \textbf{DET} & \textbf{PROPN} & \textbf{NOUN}\\ 
\hline \hline
Baseline & 32.33 & 32.14 & 33.02 & 32.93\\ 
\ \ \ \ +HAN(\textit{k} = {+1}) & 32.93$^{\spadesuit}$ & 32.68$^{\spadesuit}$ & 33.98$^{\spadesuit}$ & 33.76$^{\spadesuit}$ \\ 
\hline
& \multicolumn{4}{|c}{\textbf{German$\rightarrow$English}}\\
\cline{2-5}
\textbf{Model} & \textbf{Cataphora} & \textbf{DET} & \textbf{PROPN} & \textbf{NOUN}\\ 
\hline \hline
Baseline & 36.91 & 36.35 & 38.81 & 38.84\\
\ \ \ \ +HAN(\textit{k} = {+1}) & 37.68$^{\spadesuit}$ & 37.19$^{\spadesuit}$ & 39.51 & 39.45\\
\hline \hline
& \multicolumn{4}{c}{\textbf{English$\rightarrow$Portuguese}} \\
\cline{2-5}
\textbf{Model} & \textbf{Cataphora} & \textbf{DET} & \textbf{PROPN} & \textbf{NOUN}\\ 
\hline \hline
Baseline & 36.29 & 35.91 & 37.91 & 37.60 \\
\ \ \ \ +HAN(\textit{k} = {+1}) & 37.08$^{\spadesuit}$ & 36.70$^{\spadesuit}$ & 38.49 & 38.19 \\
\hline
& \multicolumn{4}{|c}{\textbf{Portuguese$\rightarrow$English}}\\
\cline{2-5}
\textbf{Model} & \textbf{Cataphora} & \textbf{DET} & \textbf{PROPN} & \textbf{NOUN}\\ 
\hline \hline
Baseline & 40.74 & 40.12 & 42.77 & 42.63\\
\ \ \ \ +HAN(\textit{k} = {+1}) & 41.63$^{\spadesuit}$ & 41.06$^{\spadesuit}$ & 43.60$^{\spadesuit}$ & 43.42$^{\spadesuit}$\\
\end{tabular}}
}
\caption{BLEU on the cataphora test set and its subsets for the Transformer and Transformer-HAN (\textit{k} = +1). $\spadesuit$: Statistically significantly better than the baseline.
}
\label{table:auxresults}
\vspace{-2mm}
\end{table}

\paragraph{Ablation Study}
We would like to investigate whether a context-aware NMT model trained on a wider context could perform well even if we do not have access to the same amount of context at decoding. 
We thus perform an ablation study for English$\rightarrow$German using the HAN model trained with two previous and next sentences as context and decoded with variant degrees of context.

From Table~\ref{table:ablation},  we note that reducing the amount of context at decoding time does not have adverse effect on the model's performance. However, when no context is used, there is a statistically significant drop in BLEU, 
while APT and F1-scores are equivalent to that of the baseline. This suggests that the model does rely on the context to achieve the improvement in pronoun translation. {Further, we find that the future context is just as beneficial as the past context in improving general translation performance.}


\paragraph{Cataphora-Focused Test Suite}
To gauge if the improvements in Table~\ref{table:mainresults} for the HAN (\textit{k} = +1) model are coming from the correct translation of cataphoric pronouns, we perform an evaluation on a cataphoric pronoun test suite constructed from the held-out set mentioned earlier in \S~\ref{sec:data}. To this end, we apply \texttt{NeuralCoref} over the English side to extract sentence-pairs which have a cataphoric pronoun in one sentence and the postcedent in the next sentence
. This is further segmented into subsets based on the part-of-speech of the postcedent, that is, determiner (DET), proper noun (PROPN) or all nouns (NOUN) (more details in the appendix).\footnote{{We note that there may be some overlap between the three pronoun subsets as a test sentence may contain more than one type of pronoun.}} 

From Table~\ref{table:auxresults}, {we observe HAN (\textit{k} = +1) to outperform the baseline for all language-pairs when evaluated on the cataphora test suite. In general, we observe greater improvements in BLEU when translating to English, which we attribute to the simplification of cross-lingual pronoun rules when translating from German or Portuguese to English.\footnote{It should be noted that the cataphora test set is extracted based on the existence of cataphoric-pairs in the English-side, which may have biased the evaluation when English was in the target.} We also observe fairly similar gains in BLEU across the different pronoun subsets, which we hypothesise to be due to potential overlap in test sentences between different subsets. Nevertheless,} we note optimum translation quality over the noun subsets (PROPN and NOUN), while seeing the greatest percentage improvement on the DET subset. For the latter, we surmise that the model is able to more easily link pronouns in a sentence to subjects prefixed with possessive determiners, for example, ``his son" or ``their child". 


{We also perform an auxiliary evaluation for Transformer-HAN (\textit{k} = {-1}) trained with the previous sentence as context on the cataphora test suite and find that the BLEU improvements still hold. Thus, we conclude that Transformer-HAN (a context-aware NMT model) is able to make better use of coreference information to improve translation of pronouns (detailed results in Appendix~\ref{section:app-cataphora}).} 

\paragraph{Qualitative Analysis}

\begin{figure}[t!]
 \centering
      \includegraphics[width=1\linewidth]{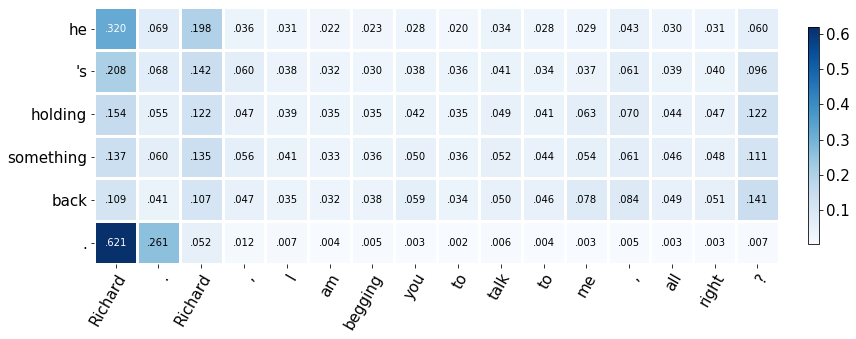}
  \caption{Example attention map between source (\textit{y}-axis) and context (\textit{x}-axis). The source pronoun \textit{he} correctly attends to the postcedents \textit{Richard} in the context.}
 \label{fig:attsrc-ctx}
\end{figure}

We 
analyse the distribution of attention to the context sentence for a few test cases.\footnote{Attention is average of the per-head attention weights.} 
Figure~\ref{fig:attsrc-ctx} shows an example in which a source pronoun \textit{he} attends to its corresponding postcedent in context. This is consistent with our hypothesis that the HAN (\textit{k} = +1) is capable of exploiting contextual information for the resolution of cataphoric pronouns.

\section{Conclusions}
In this paper, we have investigated the use of future context for NMT and particularly for pronoun translation. While previous works have focused on the use of past context, we demonstrate through rigorous experiments that using future context does not deteriorate translation performance over a baseline. Further, it shows comparable and in some cases better performance as compared to using the previous sentence in terms of both generic and pronoun-focused evaluation. In future work, we plan to investigate translation of other discourse phenomena that may benefit from the use of future context.

\section*{Acknowledgments}

The authors are grateful to the anonymous reviewers for their helpful comments and feedback {and to George Foster for fruitful discussions}.
%
%
This work is supported by a Google Faculty Research Award to G.H.
It is further supported by the Multi-modal Australian ScienceS Imaging and Visualisation Environment (MASSIVE) (\url{www.massive.org.au}).

\bibliography{acl2020}
\bibliographystyle{acl_natbib}

\appendix

\section{Experiments}
\subsection{Model Configuration}\label{section:app-modelconfig}
We use similar configuration as the Transformer-base model \cite{Vaswani:17} except that we reduce the number of layers in the encoder and decoder stack to 4 following \newcite{Maruf:2019}. For training, we use the default Adam optimiser \cite{Kingma:14} with an initial learning rate of 0.0001 and employ early stopping. 

\begin{figure}[ht]
\centering
%
\subcaptionbox{Europarl}
{
\begin{tikzpicture}
    \begin{axis}[
            ybar stacked,
            x=0.4cm,
            bar width=0.21cm,
            height=.53\linewidth,
            legend style={at={(0.64,1)},
                anchor=north,legend columns=-1},
            symbolic x coords={0, 1, 2, 3, 4, 5, 6, 7, 8, 9, 10},
            xtick={0, 1, 2, 3, 4, 5, 6, 7, 8, 9, 10},
	every node near coord/.append style={font=\scriptsize},
            ymin=0,ymax=30,
            ylabel={\%age of occurences},
        ]
        \addplot[fill=violet!60] coordinates{
        (1, 18.1) (2, 1.3) (3, 0.4) (4, 0.2) (5, 0.1) (6, 0.05) (7, 0.03) (8, 0.01) (9, 0.008) (10, 0.003)};
        \addplot[fill=orange!80] coordinates{
        (1, 3.7) (2, 0.6) (3, 0.2) (4, 0.1) (5, 0.07) (6, 0.04) (7, 0.02) (8, 0.01) (9, 0.007) (10, 0.004)};
        \legend{\scriptsize{Anaphora}, \scriptsize{Cataphora}}
    \end{axis}
\end{tikzpicture}
}
\subcaptionbox{TED Talks}
{
\begin{tikzpicture}
    \begin{axis}[
            ybar stacked,
            x=0.4cm,
            bar width=0.21cm,
            height=.53\linewidth,
            legend style={at={(0.64,1)},
                anchor=north,legend columns=-1},
            symbolic x coords={0, 1, 2, 3, 4, 5, 6, 7, 8, 9, 10},
            xtick={0, 1, 2, 3, 4, 5, 6, 7, 8, 9, 10},
	every node near coord/.append style={font=\scriptsize},
            ymin=0,ymax=30,
            ylabel={\%age of occurences},
        ]
        \addplot[fill=violet!60] coordinates{
        (1, 23.2) (2, 3) (3, 1.2) (4, 0.6) (5, 0.34) (6, 0.29) (7, 0.15) (8, 0.13) (9, 0.08) (10, 0.05)};
        \addplot[fill=orange!80] coordinates{
        (1, 4.6) (2, 1.14) (3, 0.5) (4, 0.28) (5, 0.2) (6, 0.18) (7, 0.11) (8, 0.09) (9, 0.05) (10, 0.05)};
        \legend{\scriptsize{Anaphora}, \scriptsize{Cataphora}}
    \end{axis}
\end{tikzpicture}
}
\caption{Plots showing proportion of intersentential English pronouns versus size of coreference resolution window for Europarl and TED Talks corpora.}\label{fig:app-corpusanalysis}
\end{figure}
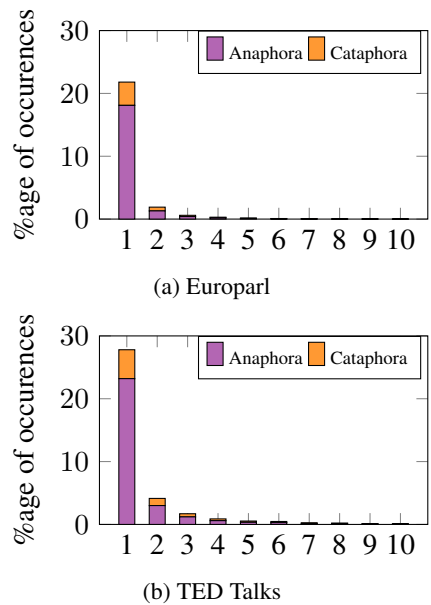

\subsection{English$\rightarrow$Russian Experiments}
\label{section:app-enru}
We wanted to compare the two variants of Transformer-HAN with \textit{k} = +1 and \textit{k} = -1 in the same experimental setting as done by \newcite{Voita:18}. The data they made available only contains the previous context sentence. Thus, we extract training, development and test sets following the procedure in this work but of roughly the same size as \newcite{Voita:18} for a fair comparison of the two context settings. While they extract their test set as a random sample of sentences, we extract from a random sample of documents, resulting in a test set which has document-level continuity between sentences. The pre-processing pipeline is the same as the one used for our English-German and English-Portuguese experiments except that we perform lowercasing (instead of truecasing) and learn separate BPE codes for source and target languages following \newcite{Voita:18}. We also evaluate the models trained with our training set on the test set provided by \newcite{Voita:18} after removing the sentences overlapping with our train and dev sets (corpora statistics are in Table~\ref{table:app-corpus}).

\setlength{\tabcolsep}{1.5pt}
\begin{table}[b]
\centering
{\small
\begin{tabular}{l||c|c}
\textbf{Origin} & 
\textbf{\#Sentences} & \textbf{Document length}\\
\hline \hline
\footnotesize{\newcite{Voita:18}} & 2M/10K/10K & -\\
\hline
Ours & 2M/11K/10K & 606.3/620.6/631.6 \\
Our, \footnotesize{\newcite{Voita:18}}$^\star$ & 2M/11K/7.3K & 606.3/620.6/-\\
\end{tabular}
}
\caption{Train/dev/test statistics for English-Russian: number of sentences (K: thousands, M: millions), and average document length (in sentences). The first row mentions statistics of data used by \newcite{Voita:18}, the second row mentions statistics of data we extracted, and the third row mentions the data statistics for our train/dev sets and \newcite{Voita:18}'s test after removing overlap (referred as \newcite{Voita:18}$^\star$).}
\label{table:app-corpus}
\end{table}

\paragraph{Results}
Table~\ref{table:app-enru} indicates that the model trained with the next sentence as context not only statistically significantly outperforms the Transformer baseline (+0.9 BLEU) but also demonstrates comparable performance to the HAN model trained with the previous sentence. This finding is consistent with our main results. We also evaluate the model trained with our training data on \newcite{Voita:18}$^\star$ test set and report almost four points jump in the absolute BLEU score for both the baseline and the context-dependent model.\footnote{The BLEU score for the baseline on \newcite{Voita:18}$^\star$ is less than the one reported in their original work because of the reduced size of the test set and the different training set.} In addition, we note that for their test set, the HAN (\textit{k} = -1) has greater percentage improvement over the baseline (4\%) than what they report for their context-aware model (2.3\%). 

\setlength{\tabcolsep}{0.3pt}
\begin{table}[t]
\centering
{\small
\begin{tabular}{l||c| c c}
\textbf{Data Setting} & \textbf{Baseline} & \textbf{HAN(\textit{k} = {+1})} & \textbf{HAN(\textit{k} = {-1})} \\
\hline \hline
Ours & 23.35 & \textbf{24.25} & 24.18\\
Our, \footnotesize{\newcite{Voita:18}}$^\star$ & 27.15 & - & \textbf{28.23}\\
\end{tabular}
}
\caption{BLEU on tokenised lowercased text for the Transformer baseline and Transformer-HAN  with following sentence (\textit{k} = +1) and previous sentence (\textit{k} = -1) for English$\rightarrow$Russian. All reported results for the HAN variants are statistically significantly better than the baseline.} 
\label{table:app-enru}
\end{table}

\subsection{Cataphora-Focused Test Suite}
\label{section:app-cataphora}
We segment the cataphora test set into the following subsets based on the part-of-speech of the postcedent being referred to:
\begin{itemize}
  \item \textbf{DET} Postcedents prefixed with possessive determiners, e.g., \textit{his son} or \textit{their child}.
  \item \textbf{PROPN} Postcedents which are proper nouns, i.e., named entities.
  \item \textbf{NOUN} Postcedents which are nouns, including proper nouns and common nouns, such as \textit{boy} or \textit{child}.
\end{itemize}


\subsubsection{Results for HAN (\textit{k} = -1)}
\label{section:app-han-anaphoric}
{We evaluate Transformer-HAN (\textit{k} = -1) enriched with anaphoric context on the cataphora test set (Table~\ref{table:app-cataphoraprev}) to determine if this context-aware model is making use of coreference information to improve the overall translation quality (in BLEU). We find that HAN (\textit{k} = +1) performs better than HAN (\textit{k} = -1) when English is in the target-side, which we hypothesise to be because of the extraction of the cataphora test suite from the English-side. However, when English is in the source-side, both models perform comparably showing that the Tranformer-HAN (a context-aware NMT model) is able to make better use of coreference information to improve translation of pronouns.}
\setlength{\tabcolsep}{3.5pt}
\begin{table}[ht]
\begin{center}
{\small
\begin{tabular}{l||c| c}
\textbf{Lang. Pair} & \textbf{Baseline} & \textbf{HAN(\textit{k} = {-1})} 
\\
\hline \hline
English$\rightarrow$German & 32.33 & 32.94\\ 
German$\rightarrow$English & 36.91 & 37.23\\
English$\rightarrow$Portuguese & 36.29 & 37.24\\
Portuguese$\rightarrow$English & 40.74 & 41.25\\
\end{tabular}
}
\end{center}
\caption{BLEU on the cataphora test set for the Transformer and Transformer-HAN (\textit{k} = -1). All results for HAN (\textit{k} = -1) are statistically significantly better than the baseline.
}
\label{table:app-cataphoraprev}
\vspace{-2mm}
\end{table}

\subsection{Pronoun Sets}\label{section:app-pronsets}
\setlength{\tabcolsep}{1.25pt}
\begin{table}[htbp]
\centering
{\small
\begin{tabular}{l|l}
\textbf{Language} & \textbf{Pronouns}\\
\hline
English & i, me, my, mine, myself, we, us, our, ours, \\
& ourselves, you, your, yours, yourself, yourselves, \\
& he, his, him, himself, she, her, hers, herself, it, \\
& its, itself, they, them, their, themselves, that, this,\\
& these, those, what, whatever, which, whichever, \\
& who, whoever, whom, whose\\
\hline
German & ich, du, er, sie, es, wir, mich, dich, sich, ihn,\\
& uns, euch, mir, dir, ihm, ihr, ihre, ihrer, ihnen,\\
& meiner, mein, meine, deiner, dein, seiner, sein,\\
& seine, unser, unsere, euer, euere, denen, dessen,\\
& deren, meinen, meinem, deinen, deinem, deines, \\
& unserer, unseren, unseres, unserem, ihrem, ihres,\\
& seinen, seinem, seines\\
\hline
{Portuguese} & eu, n{\'o}s, tu, voc{\^e}, voc{\^e}s, ele, ela, eles, elas,\\
& me, te, nos, vos, o, lo, no, a, la, na, lhe, se, os, \\
& los, as, las, nas, lhes, mim, ti, si, meu, teu, seu, \\
& nosso,vosso, minha, tua, sua, nossa, vossa, meus \\
& teus, seus, nossos, vossos, minhas, tuas, suas \\
& nossas, vossas, dele, dela, deles, delas, quem \\
& que, qual, quais, cujo, cujos, cuja, cujas, onde \\
\end{tabular}
}
\caption{Pronoun sets used in our pronoun-focused automatic evaluation.}
\label{table:app-pronlist}
\end{table}

\end{document}